\documentclass[sigconf]{acmart}

\AtBeginDocument{%
  }

\usepackage{subcaption}
\usepackage{multirow}
\usepackage{amsmath}
\usepackage{hyperref}
\usepackage{balance}

\copyrightyear{2022}
\acmYear{2022}
\setcopyright{acmcopyright}\acmConference[MM '22]{Proceedings of the 30th ACM
International Conference on Multimedia}{October 10--14, 2022}{Lisboa, Portugal}
\acmBooktitle{Proceedings of the 30th ACM International Conference on Multimedia
(MM '22), October 10--14, 2022, Lisboa, Portugal}
\acmPrice{15.00}
\acmDOI{10.1145/3503161.3548317}
\acmISBN{978-1-4503-9203-7/22/10}

\begin{document}

\title{Exploiting Transformation Invariance and Equivariance for Self-supervised Sound Localisation}


\author{Jinxiang Liu}
\orcid{0000-0003-2583-8881}
\affiliation{%
  \institution{Cooperative Medianet Innovation Center\\ Shanghai Jiao Tong University}
  \country{China}
}
\email{jinxliu@sjtu.edu.cn}

\author{Chen Ju}
\affiliation{%
  \institution{Cooperative Medianet Innovation Center\\Shanghai Jiao Tong University}
  \country{China}
}
\email{ju_chen@sjtu.edu.cn}

\author{Weidi Xie}
\authornote{Corresponding author.}
\affiliation{%
  \institution{$^1$Cooperative Medianet Innovation Center\\ Shanghai Jiao Tong University}
  \country{China}
}
\affiliation{\institution{$^2$Shanghai AI Laboratory}\country{China}}
\email{weidi@sjtu.edu.cn}

\author{Ya Zhang}
\authornotemark[1]
\affiliation{%
  \institution{$^1$Cooperative Medianet Innovation Center\\ Shanghai Jiao Tong University}
  \country{China}
}
\affiliation{\institution{$^2$Shanghai AI Laboratory}\country{China}}
\email{ya_zhang@sjtu.edu.cn}


\renewcommand{\shortauthors}{Jinxiang Liu, Chen Ju, Weidi Xie, \& Ya Zhang}

\newcommand{\etal}{\textit{et al}.}
\newcommand{\xmark}{\ding{55}}%
\begin{abstract}
We present a simple yet effective self-supervised framework for audio-visual representation learning,  to localize the sound source in videos. To understand what enables to learn useful representations, we systematically investigate the effects of data augmentations, and reveal that (1) composition of data augmentations plays a critical role, {\em i.e.}~explicitly encouraging the audio-visual representations to be invariant to various transformations~({\em transformation invariance}); (2) enforcing geometric consistency substantially improves the quality of learned representations, {\em i.e.}~the detected sound source should follow the same transformation applied on input video frames~({\em transformation equivariance}). Extensive experiments demonstrate that our model significantly outperforms previous methods on two sound localization benchmarks, namely, Flickr-SoundNet and VGG-Sound. Additionally, we also evaluate audio retrieval and cross-modal retrieval tasks. In both cases, our self-supervised models demonstrate superior retrieval performances, even competitive with the supervised approach in audio retrieval. This reveals the proposed framework learns strong multi-modal representations that are beneficial to sound localisation and generalization to further applications.
\textit{The project page is \href{https://jinxiang-liu.github.io/SSL-TIE}{https://jinxiang-liu.github.io/SSL-TIE/} 
}.
\end{abstract}

\ccsdesc[500]{Information systems~Multimedia information systems}
\ccsdesc[500]{Computing methodologies~Computer vision}
\ccsdesc[500]{Computing methodologies~Unsupervised learning}

\keywords{self-supervised representation learning, sound localisation}
\begin{teaserfigure}
\vspace{-0.1cm}
{
\centering
\begin{subfigure}[b]{ \textwidth}
    \centering
    \includegraphics[width=\textwidth]{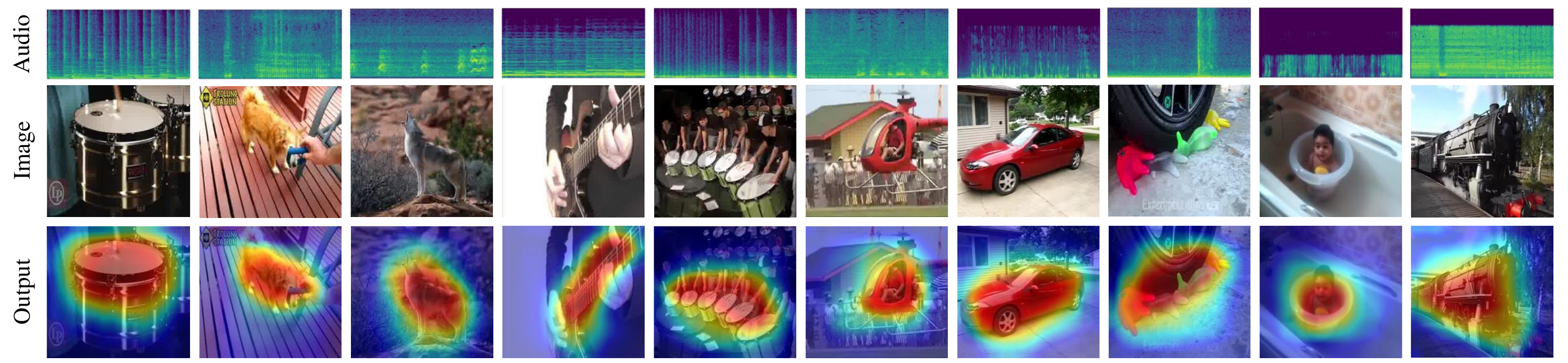}
    \vspace{-0.5cm}
    \caption{Visual Sound Localisation }
    \vspace{0.2cm}
    \end{subfigure}
    \hfill
\begin{subfigure}[b]{0.49\textwidth}
    \centering
    \includegraphics[width=\textwidth]{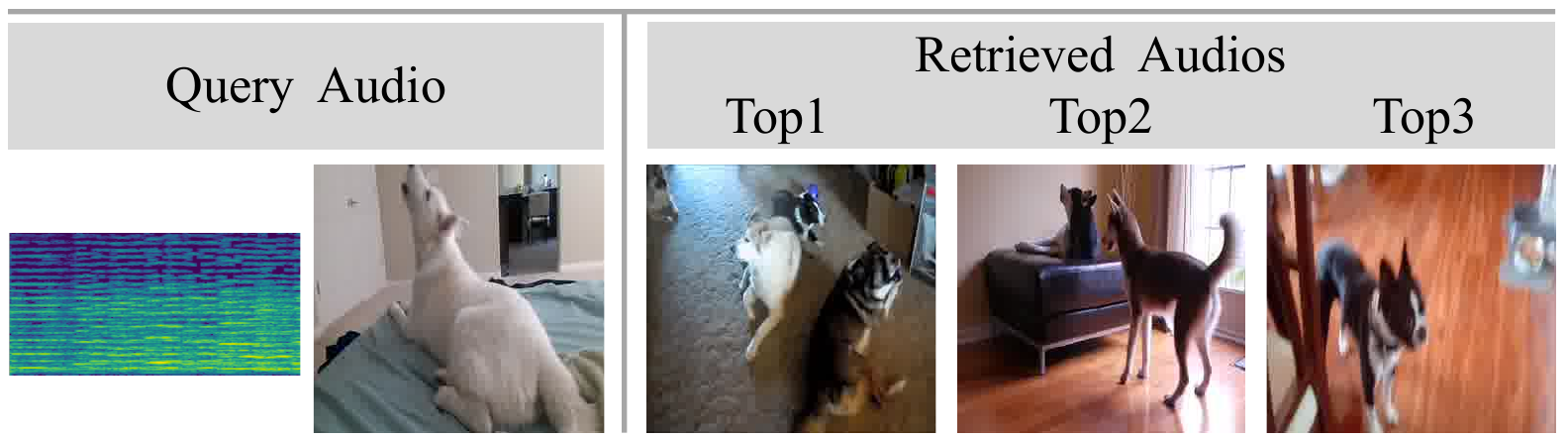}
    \vspace{-0.5cm}
    \caption{Audio Retrieval }
    \end{subfigure}
    \hfill
\begin{subfigure}[b]{0.49\textwidth}
    \centering
    \includegraphics[width=\textwidth] {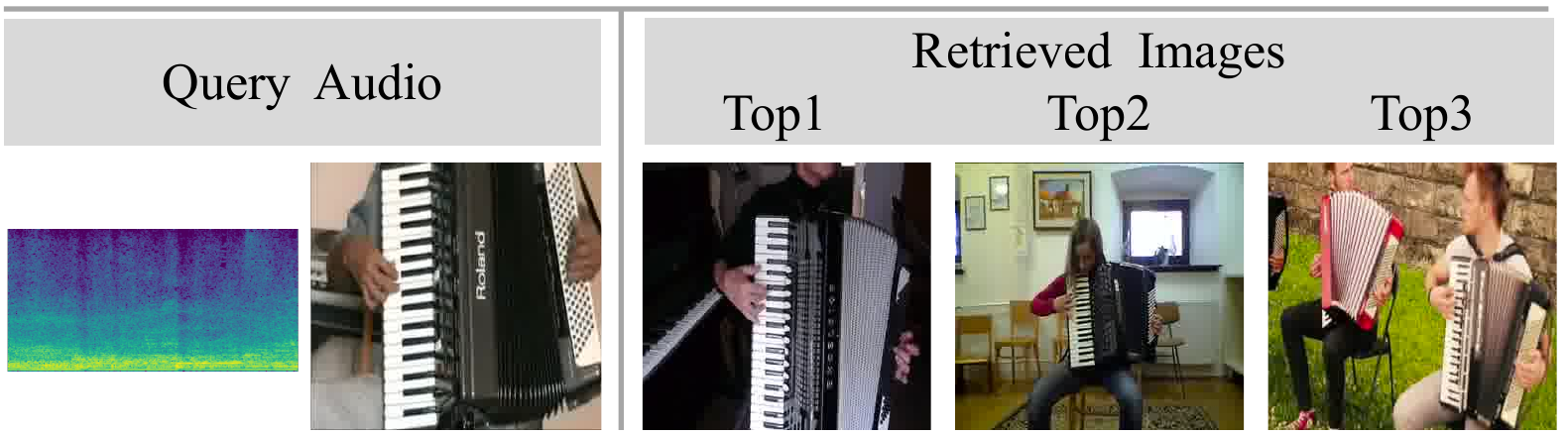}
    \vspace{-0.5cm}
    \caption{Cross-modal Retrieval}
    \end{subfigure}
\vspace{-0.2cm}
\caption{Visualized results of our framework which learns powerful multi-modal representations for various applications.
(a) visual sound localisation: highlights the salient object by its emitted sound. 
(b) audio retrieval: discovers semantic-identical audios to the query audio. 
(c) cross-modal retrieval: discovers semantic-identical images to the query audio.}
\label{fig:teaser}
}
\end{teaserfigure}

\maketitle

\section{Introduction}

When looking around the world, we can effortlessly perceive the scene from multi-sensory signals, for example, whenever there is sound of dog barking, we would also expect to see a dog somewhere in the scene.
A full understanding of the scene should thus include the interactions between the visual appearance and acoustic characteristics.
In the recent literature, researchers have initiated research on various audio-visual tasks, including audio-visual sound separation~\cite{gao2019co,gao2018learning,gan2020music,gao2021visualvoice,xu2019recursive,av_iclr21_AudioScope,zhao2018sound,zhao2019sound}, visual sound source localisation~\cite{senocak2019learning,qian2020multiple,hu2020discriminative,chen2021localizing,hu2019deep,song2022sspl,hu2021class,lin2021unsupervised} and audio-visual video understanding~\cite{wang2020makes,gao2020listen,xiao2020audiovisual,kazakos2019epic,tian2020unified,lin2019dual,lee2020cross}.
In this paper, we focus on the task of visual sound source localisation, with the goal to highlight the salient object by its emitted sound in a given video frame.
To avoid the laborious annotations, we here consider a self-supervised setting, which only requires raw videos as the training data, {\em i.e.} without using any extra human annotations whatsoever. 


Generally speaking, the main challenge of visual sound localisation is to learn joint embeddings for visual and audial signals. 
To this end, various attempts have been made in early works. 
\cite{arandjelovic2018objects,senocak2019learning} train classification models to predict whether audio and video frame are corresponding or not. And the localisation representation is obtained by computing similarity between audio and image representations, revealing the location of sounding objects;
Qian~\etal~\cite{qian2020multiple} also learn audio and visual representations with the classification model to localise sounding objects, they leverage the pre-trained classifiers to aggregate more audio-image pairs of the same semantics by comparing their category labels.
More recent work~\cite{chen2021localizing} has tried to explicitly mine the sounding regions automatically through differentiable thresholding, and then self-train the model with the InfoNCE loss~\cite{van2018representation}.
Despite tremendous progress has been made, 
previous visual sound source localisation approaches have always neglected the important role of \textit{data augmentations}, 
which has shown to be essential in self-supervised representation learning~\cite{chen2020simple,he2020momentum,grill2020bootstrap,chen2021exploring}.

Herein, we introduce a simple self-supervised framework to explore the efficacy of data transformation.
Specifically, we exploit Siamese networks to process two different augmentations of the audio-visual pairs, and train the model with contrastive learning and geometrical consistency regularization, 
{\em i.e.}~encouraging the audio-visual correspondence to be \textit{invariant} to various transformations, 
while enforcing the localised sound source to be \textit{equivariant} to geometric transformations.
To validate the effectiveness of the proposed idea, 
we experiment with two prevalent audio-visual localisation benchmarks, namely, Flickr-SoundNet and VGG Sound-Source. 
Under the self-supervised setting, 
our approach demonstrates state-of-the-art performance, 
surpassing existing approaches by a large margin, 
even with less than $1/14$ training data, 
thus being more data-efficient. 
Additionally, we also measure the quality of learned representations by two different retrieval tasks, {\em i.e.}~audio retrieval and audio image cross-modal retrieval, which demonstrates the powerful representation learning ability of the proposed self-supervised framework.

To summarise, our main contributions are three-fold:
(i) We introduce a simple self-supervised framework to explore the efficacy of data transformation for visual sound localisation, concretely, we optimise a Siamese network with contrastive learning and geometrical consistency;
(ii) We conduct extensive experiments and thorough ablations to validate the necessity of different augmentations, and demonstrate state-of-the-art performance on two standard sound localisation benchmarks while being more data-efficient;
(iii) We initiate two audio retrieval benchmarks based on VGGSound, and demonstrate the usefulness of learned representations, {\em e.g.}~audio retrieval and cross-modal retrieval.
In both cases, our method shows impressive retrieval performances.
Codes and dataset splits will be publicly released to facilitate future research.

\begin{figure*}[t]
\centering
\vspace{0.2cm}
\includegraphics[width=.98\linewidth]{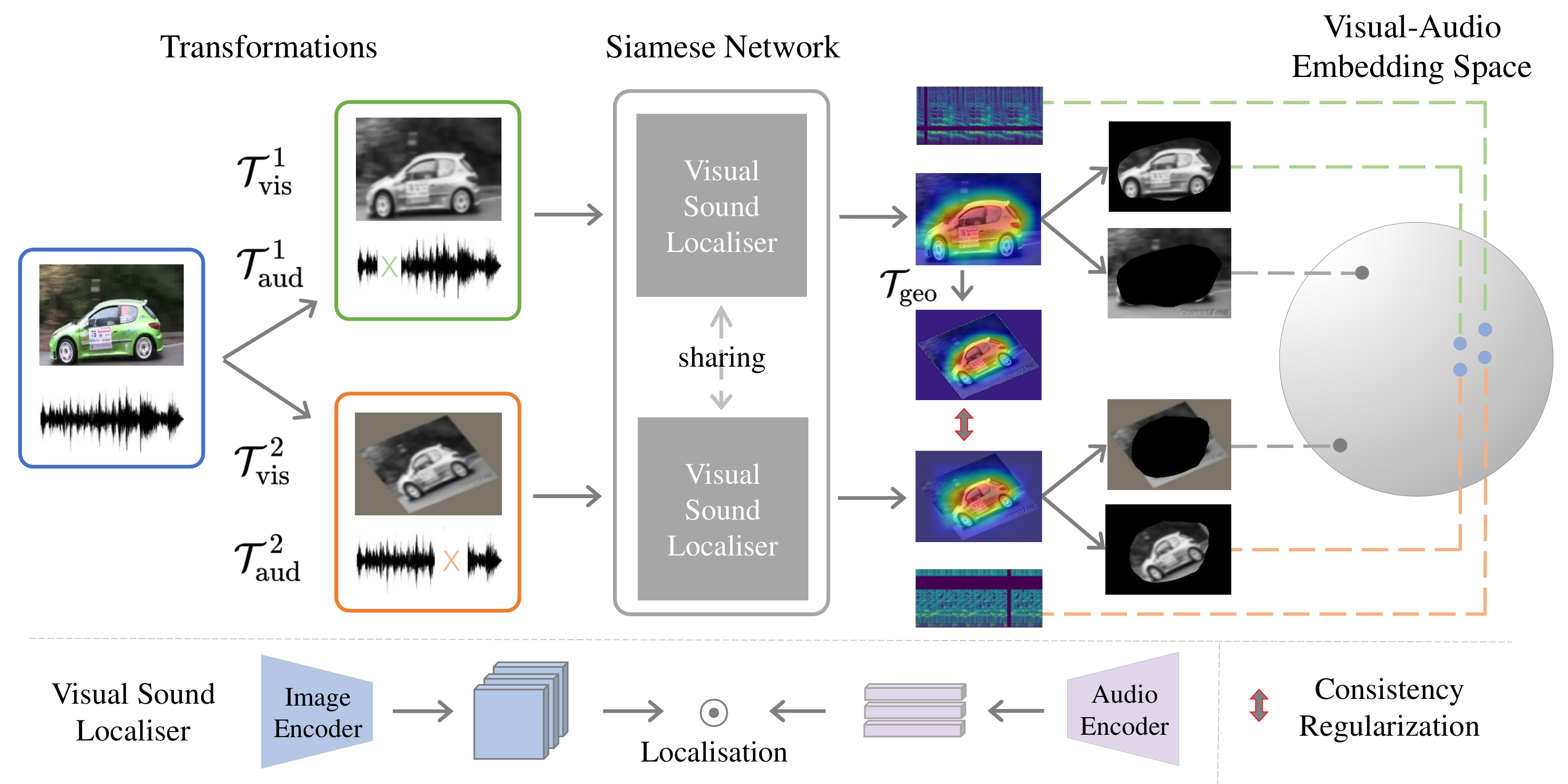}
\caption{Framework Overview. We exploit a Siamese network, with two identical branches, each branch consists of an image encoder and an audio encoder. For the one branch, we perform transformations $\mathcal{T}_{\text{vis}}^1+\mathcal{T}_{\text{aud}}^1$, while for the other branch, we use transformations $\mathcal{T}_{\text{vis}}^2+\mathcal{T}_{\text{aud}}^2$. In this figure, $\mathcal{T}_{\text{vis}}^1$ only includes appearance transformation $\mathcal{T}_{\text{app}}$, while $\mathcal{T}_{\text{vis}}^2$ includes both appearance and geometric transformations $\mathcal{T}_{\text{app}}+\mathcal{T}_{\text{geo}}$.
Both audio transformations are $\mathcal{T}_{aud}$. The framework is optimised by encouraging the audio-visual representation to be invariant to $\mathcal{T}_{app}$ and $\mathcal{T}_{geo}$, while being equivalent to $\mathcal{T}_{geo}$.}
\vspace{0.2cm}
\label{fig:framework}
\end{figure*}

\section{Related work}
In this section, we first review previous work on audio-visual sound source localisation, especially on the self-supervised methods; we then describe the research on self-supervised representation learning with Siamese networks; finally, we summarize the literature regarding transformation equivariance.

\subsection{Self-supervised Sound Localisation}
Audio-visual sound source localisation aims to localise the object region that corresponds to the acoustic sound in a given video frame.
Early approaches have exploited the statistical models to maximize the mutual information between different modalities~\cite{hershey1999audio,fisher2000learning}. 
Recently, deep neural networks have been adopted for representation learning,
by leveraging the innate synchronization between audio and video, 
for example, SSMF~\cite{owens2018audio} and AVTS~\cite{korbar2018cooperative} deploy networks to predict whether visual contents and audio are temporally aligned or not, 
then the sounding objects can be discovered through Class Activation Mapping (CAM)~\cite{zhou2016learning}.
Senocak~\etal~\cite{senocak2018learning} develop a foreground attention mechanism with the triplet loss~\cite{hoffer2015deep},
where the attention map is computed by the inner dot product between sound and visual context.
Qian~\etal~\cite{qian2020multiple} propose a two-stage framework for multiple-object sound localization, 
they first leverage the pre-trained classifiers to obtain pseudo category labels and then align the multi-modal features.
Such pipeline is not end-to-end trainable, thus may hinder the performance. 

Recently, contrastive learning with infoNCE loss~\cite{van2018representation} has shown great success in self-supervised representation learning~\cite{chen2020simple,he2020momentum}.
The methods including SimCLR~\cite{chen2020simple} and MoCo~\cite{he2020momentum}  construct various augmentations of the same samples as positive pairs, 
while the augmentations of other samples as the negatives, 
resembling an instance discrimination task.
Inspired by this, Chen~\etal~\cite{chen2021localizing} introduce the infoNCE contrastive learning to sound source localisation, 
where they treat the responses of the sounding object within the foreground image with its corresponding audio as positive,
while the responses of background image with audio and the responses of mismatched image-audio pairs as negatives.
However, the authors ignore the importance of image data augmentations, 
which have proven to be critical in the self-supervised instance discrimination models \cite{he2020momentum,chen2020simple,grill2020bootstrap,chen2021exploring}.
In this paper, we intend to fill this gap by exploring various data transformations.

\subsection{Siamese Network}
The Siamese network, which consists of two or more identical sub-networks, 
is typically used to compare the similarity between predictions brought by different entities. 
It is prevalent to solve many problems, 
including face verification~\cite{taigman2014deepface}, 
visual tracking~\cite{bertinetto2016fully,leal2016learning}, 
one-shot object recognition~\cite{koch2015siamese}, 
and recommendation~\cite{maheshwary2018matching}. 
More recently, the Siamese network has been widely adopted for self-supervised representation representation learning~\cite{he2020momentum,chen2020simple,grill2020bootstrap,chen2021exploring}.
Concretely, the contrastive learning methods, such as SimCLR~\cite{chen2020simple} and MoCo~\cite{he2020momentum}, 
aim to attract two augmented views of the same image while push away views from different image samples with the InfoNCE loss, thus resembling an instance discrimination loss.
BYOL~\cite{grill2020bootstrap}, SimSiam~\cite{chen2021exploring} and ContrastiveCrop~\cite{peng2022crafting} feed two branches of Siamese networks with different augmentations of the same image sample, 
and they utilize one branch to predict the output of the other.
To the best of our knowledge, this is the first exploration to leverage the Siamese networks for sound localisation based on the contrastive learning.

\subsection{Equivariant Transformation}
Equivariant transformation refers that the predictions from a model are equivariant to the transformations applied to the input images.
It is a popular technique in many problem which requires spatial prediction such as unsupervised landmark localisation~\cite{thewlis2017unsupervised,thewlis2019unsupervised}.
The assumption~\cite{thewlis2017unsupervised,thewlis2019unsupervised} is that the learned landmark should be consistent with the visual effects of image deformations such as viewpoint change or object deformation.
The transformation equivariance is also prevalent for some problems in semi-supervised settings including landmark localisation~\cite{honari2018improving,moskvyak2021semisupervised}, image segmentation~\cite{wang2020self}, image-to-image translation~\cite{mustafa2020transformation}. 
The common approach of ~\cite{honari2018improving,moskvyak2021semisupervised,wang2020self,mustafa2020transformation} is to train the models with the labelled data and enforce the predictions for the unlabelled data to be equivariant to the transformations applied on them.
In this paper, we exploit the transformation equivariance property by integrating it into the proposed unified self-supervised framework for sound localization.

\section{Method}
In this paper, we consider the self-supervised audio-visual representation learning, to localise the sound source in the video frames. In Section~\ref{sec:pre-contra}, we first introduce the general problem scenario; In Section~\ref{sec:Visual Sound Localisation}, we introduce the proposed Siamese framework~(Figure~\ref{fig:framework}), and describe different data transformations for both audio and visual signals; Lastly, in Section~\ref{sec:trans-inv and trans-equ}, we propose the essential transformation invariance and equivariance, and also summarize the training objectives for joint model optimisation.

\subsection{Problem Scenario}
\label{sec:pre-contra}
In visual sound localisation, we are given a set of raw videos $\mathcal{X}=\left\{\left(I_{1}, A_{1}\right),\left(I_{2}, A_{2}\right), \cdots, \left(I_{N}, A_{N}\right)\right\}$, where $I_{i} \in \mathbb{R}^{3 \times H_{v} \times W_{v}}$ refers to the central frame of \textit{i}-th video, $A_{i} \in \mathbb{R}^{1 \times H_{a} \times W_{a}}$ denotes its corresponding audio spectrogram, $H_v, W_v$ and $H_a, W_a$ are the spatial resolutions of two modalities respectively. 
The goal is to learn a visual localisation network that takes the audio-visual pair as inputs and outputs the localisation map for sounding object:
\begin{align}
    \Phi_{\text{loc}}(I_i, A_i; \Theta) =
    \mathbf{M}_{\text{loc}} \in \{0, 1\}^{H_v \times W_v}
\end{align}
where $\Theta$ represents the learnable parameters, and $\mathbf{M}_{\text{loc}}$ refers to a binary  segmentation mask, with $1$ denoting the visual location of objects that emit the sound.

\subsection{Visual Sound Localisation}
\label{sec:Visual Sound Localisation}
In order to learn the joint audio-visual embedding, we here exploit a Siamese network with two identical branches.
As shown in Figure~\ref{fig:framework}, each branch is consisted of an image encoder~($f_v(,:\theta _v)$) and an audio encoder~($f_a(,:\theta _a)$), and the embeddings of two modalities can be computed as follows:
\begin{equation}
\begin{aligned}
    v & = f_v(\mathcal{T}_{\text{vis}}(I), \theta _v), \quad v\in \mathbb{R}^{c\times h \times w}  \\
    a & = f_a(\mathcal{T}_{\text{aud}}(A), \theta _a), \quad a\in \mathbb{R}^{c},
\end{aligned}
\end{equation}
where $\mathcal{T}_{\text{vis}}$ and $\mathcal{T}_{\text{aud}}$ refer to the augmentations imposed on visual frames and audio spectrograms, respectively. $h$, $w$ refer to the visual spatial resolution of the visual feature map, and $c$ denotes the dimension of the encoded audio vector.

To localise the visual objects, we can thus compute the response map $S_{i \rightarrow j}$, 
by measuring the cosine distance between the audio features $a_i$ and pixel-level visual features $v_j$:
\begin{equation}
    S_{i \rightarrow j} = \frac{\left \langle a_i, v_j \right \rangle }{\left \| a_i \right \| \cdot \left \| v_j \right \| } \in \mathbb{R}^{h \times w},
\end{equation}
where $S_{i \rightarrow j}$ indicates the visual-audio activation between the $i$-th video frame and the $j$-th audio.
The final segmentation map $\mathbf{M}_{\text{loc}}$ is attained by simply thresholding $S_{i \rightarrow j}$.

\subsubsection{\bf Transformation on audio spectrogram~($\mathcal{T}_{\text{aud}}$).}
Here, before feeding audio data to the audio encoder, we pre-process the \textit{1-D} waveform to obtain \textit{2-D} mel-spectrograms, with horizontal and vertical axes representing time and frequency, respectively. 
Then, we consider two different types of audio augmentations, {\em i.e.}~spectrogram masking $\mathcal{T}_{\text{mask}}$ and audio mixing $\mathcal{T}_{\text{mix}}$.

As for spectrogram masking, we randomly replace the \textit{2-D} mel-spectrograms with zeros along two axes with random widths, that is, \textit{time masking} and \textit{frequency masking} on mel-spectrograms~\cite{park2019specaugment}. 
While for audio mixing, we aim to blend the audio samples with same semantics.
To find the semantic identical audio for each audio sample, we compute the similarity of embedding with all other audio samples in datasets and adopt the most similar one to mix.
We conduct such mixing strategy in a curriculum learning manner: the blending weights for the sampled audios are linearly increased from 0 to 0.65 as the training proceeds. Mathematically:
\begin{equation}
    A_i^{\text{mix}} = (1-\alpha)\cdot A_i + \alpha \cdot A_i^{\text{sim}}, 
\end{equation}
where $A_i^{\text{sim}}$ is the most similar audio sample of the audio $A_i$, $A_i^{\text{mix}}$ refers to the mixed audio, and $\alpha$ is the mixing coefficient, which increases linearly with the training epoch. 
In Section~\ref{sec:ablation}, we have conducted thorough experiments, showing both transformations are critical for improving sound localisation performance while preventing the model from overfitting.

\subsubsection{\bf Transformation on visual frames~($\mathcal{T}_{\text{vis}}$).}
Here, we split the image transformations into two groups: appearance transformations $\mathcal{T}_{\text{app}}$ and geometrical transformations $\mathcal{T}_{\text{geo}}$. 
$\mathcal{T}_{\text{app}}$ refers to transformations that only change the frame appearances, including color jittering, gaussian blur, and grayscale; $\mathcal{T}_{\text{geo}}$ changes the geometrical shapes and locations of the sounding objects, including cropping and resizing, rotation, horizontal flipping.
These transformations are shown to be essential for representation learning in recent visual self-supervised approaches, {\em e.g.}~SimCLR~\cite{chen2020simple}, MOCO~\cite{he2020momentum}, DINO~\cite{caron2021emerging}, etc.
We refer the readers for both audio and visual frame transformations in supplementary materials.

\subsection{Training Details}
\label{sec:trans-inv and trans-equ}
In this section, we describe how to exploit different data transformations for training visual sound localisation models.

\subsubsection{\bf Correspondence Transformation Invariance}
Though various transformations are applied on inputs, the audio-image correspondence is not altered, which means the correspondence are invariant to the transformations.
Thus we still adopt batch contrastive learning for both branches in the Siamese framework to exploit the correlation between audio-visual signals, as follows:
\begin{align}
 m_i &= \text{sigmoid}((S_{i \rightarrow i} - \epsilon) / \tau) \\
 P_{i} &= \frac{1}{\left|m_{i}\right|}\left\langle m_{i}, S_{i \rightarrow i}\right\rangle \\
 N_{i} &= \sum_{i \neq j} \frac{1}{h w} \left\langle\mathbf{1}, S_{i \rightarrow j}\right\rangle + \frac{1}{\left| 1-m_i \right |} \left\langle 1-m_{i}, S_{i \rightarrow i}\right\rangle \\
 \mathcal{L}_{\text{cl}} &= -\frac{1}{B} \sum_{i=1}^{B}\left[\log \frac{\exp \left(P_{i}\right)}{\exp \left(P_{i}\right)+\exp \left(N_{i}\right)}\right]
\end{align}

Here, $m_i \in \mathbb{R}^{h \times w}$ refers to the foreground pseudo-mask; $P_i$ denotes the positive set that is constructed by the responses within the mask; $N_i$ denotes the negative set,with two components: the responses between unpaired audio-visual signals and the responses of its own background.

\subsubsection{\bf Geometric Transformation Equivariance}
Despite the fact that $\mathcal{T}_{\text{geo}}$ on images do not change the semantic correspondences with audios, 
$\mathcal{T}_{\text{geo}}$ do change the predicted localisation result.
And ideally, the localisation results should take the same geometrical transformations as the input images experienced during the data transformation. Formally:
\begin{equation}
    \Phi_{\text{loc}}(\mathcal{T}_{\text{geo}}(I),A) = \mathcal{T}_{\text{geo}}(\Phi_{\text{loc}}(I,A)),
\end{equation}
where $\Phi_{\text{loc}}(\cdot)$ refers to the sound source localisation network, 
and $(I,A)$ denotes the frame-audio pair. 

Based on this transformation equivariance property, we implement a geometrical transformation consistency between response outputs from two branches of the Siamese framework as:
\begin{equation}
\mathcal{L}_{geo}=\left\| S^2_{i \rightarrow i} \left(\mathcal{T}_{\text{geo}}(I), A\right)-\mathcal{T}_{\text{geo}}(S^1_{i \rightarrow i}(I, A))\right\|_{2},
\end{equation}
where $S^1_{i \rightarrow i}, S^2_{i \rightarrow i}$ are response maps from the two branches of the Siamese framework, and $\left\| \cdot \right\|$ refers to the $L^2$ norm.

\subsubsection{\bf  Optimisation Objectives}
\label{sec:training}
We train the Siamese framework by jointly optimising the contrastive loss and geometrical consistency loss in a self-supervised manner,
\begin{equation}
\mathcal{L}_{\text{total}}=\mathcal{L}^1_{\text{cl}} + \mathcal{L}^2_{\text{cl}} + \lambda_{geo} \mathcal{L}_{\text{geo}},
\end{equation}
where $\mathcal{L}^1_{\text{cl}}, \mathcal{L}^2_{\text{cl}}$ refer to the contrastive loss in both branches, $\lambda_{geo}$ represents the weighs of $\mathcal{L}_{\text{geo}}$ and is set to 2.0 empirically.

\begin{table*}[t]
\caption{Ablation study on the VGG-SS test set. 
All the models are trained with VGGSound-144k dataset. The results shows that, all data transformations and optimization losses are essential. By encouraging audio-visual invariant to various transformations, while visually equivariant to geometric transformations, we achieve considerable performance gains.}
\label{tab:ablation}
\setlength{\tabcolsep}{17.5pt}
\begin{center}
\begin{tabular}{lcccccccc}
\toprule
& \multicolumn{4}{c} { Transformations } & \multicolumn{2}{c} { Objectives} & \multicolumn{2}{c} {Results} \\
\cmidrule(lr){2-5}\cmidrule(lr){6-7}\cmidrule(lr){8-9}  
Model & $\mathcal{T}_{\text{app}}$ & $\mathcal{T}_{\text{geo}}$ & $\mathcal{T}_{\text{mask}}$ & $\mathcal{T}_{\text{mix}}$ & $\mathcal{L}_{\text{cl}}$ & $\mathcal{L}_{\text{geo}}$ & cIoU & AUC \\ \midrule
A & & & & & $\checkmark$ & & 0.3292 & 0.3744 \\ \midrule
B & $\checkmark$ & & & & $\checkmark$ & & 0.3364 & 0.3721   \\
C & $\checkmark$ & $\checkmark$ & & & $\checkmark$   & & 0.3580 &  0.3847  \\
D & $\checkmark$ & $\checkmark$ & $\checkmark$ & & $\checkmark$ & & 0.3748 & 0.3887    \\
E & $\checkmark$ & $\checkmark$ & $\checkmark$  & $\checkmark$ & $\checkmark$ & & 0.3766 & 0.3937   \\
F & $\checkmark$ & $\checkmark$ & $\checkmark$ & $\checkmark$ & $\checkmark$ & $\checkmark$ & \textbf{0.3863} & \textbf{0.3965} \\
\bottomrule
\end{tabular}
\end{center}
\vspace{0.2cm}
\end{table*}


\setlength{\parindent}{0pt}
\section{Experiment}
In this section, we conduct extensive experiments for audio-visual sound localisation on two standard benchmarks and compare with existing state-of-the-art methods.
We conduct thorough ablation studies to validate the necessity of different transformations.
Additionally, based on the VGGSound dataset, we introduce two new evaluation protocols on retrievals, to further evaluate the quality of learnt audio-visual representation.

\subsection{Implementation details}
Our proposed method is implemented with PyTorch. 
The input images are all resized to $224\times 224$ spatial resolution,
with random augmentations, 
including color jitterings, {\em e.g.}~grayscale, brightness, contrast, saturation, and geometric transformations, {\em e.g.}~rotation, horizontal flipping.
For the visual and audio encoders, we here adopt the ResNet-18 backbone. 
The shape of output features from the visual encoder is $14\times 14 \times 512$,
and the shape of audio features is $1 \times 512$. 
The model is optimised with Adam using a learning rate of $10^{-4}$, and the batch size is set to $32$.
We train the model for 80 epochs on single GeForce RTX 3090 GPU.


\subsection{Visual Sound Localisation}
\subsubsection{\bf Datasets.} 
We train and evaluate on two datasets,\\[-5pt]

\par{\bf Flickr-SoundNet~\cite{aytar2016soundnet}} 
contains more than 2M unconstrained image-audio pairs. 
Following the convention~\cite{senocak2018learning}, 
we adopt a random subset with 144k image-audio sample pairs, 
and a subset with 10k random samples for training,
termed as $\textbf{Flickr-144k}$ and $\textbf{Flickr-10k}$ respectively.
For evaluation, 
we use the 250 random sampled pairs from the subset of 5k annotated sample pairs. 
In the test subsets, each audio-image pair contains an image and its corresponding 20-second-long audio, which is annotated by three different subjects for reliability. 
The annotation is the bounding box of the dominant object that emits the sound. \\[-5pt]

\par{\bf VGGSound~\cite{chen2020vggsound}} is a video dataset consisting of 200k videos, spanning 309 sounding categories. Similar to Flickr-SoundNet splits, we use the subsets $\textbf{VGGSound-144k}$ and $\textbf{VGGSound-10k}$ for training. While for evaluation, we employ the VGG-SS~\cite{chen2021localizing}, a recently-released standard testing subset from VGG-Sound dataset. 
The testing dataset contains 5k videos, each video is annotated with a bounding box for sounding objects in the center frame.

\subsubsection{\bf Metrics.}
We quantitatively measure sound source localisation performance with two metrics: 
(i) Consensus Intersection over Union (cIoU)~\cite{senocak2018learning} measures the localisation accuracy through the intersection and union between ground-truth and prediction; (ii) Area Under Curve (AUC) indicates the area under the curve of cIoU plotted by varying the threshold from 0 to 1. For both metrics, high values mean better localisation performances.

\begin{table}[t]
\setlength{\tabcolsep}{10pt}
\caption{Comparisons on the Flcikr-SoundNet test set. All the models are trained on Flickr-144k or Flickr-10k subsets. Our method significantly outperforms these competitors.} 
\label{table:quan-flickr}
\centering
\begin{tabular}{llcc}
\toprule Method & Training Set & cIoU & AUC \\
\hline
Attention~\cite{senocak2018learning}    & Flickr-10k  & 0.436 & 0.449 \\
CoarseToFine~\cite{qian2020multiple}  & Flickr-10k  & 0.522 & 0.496 \\
AVO~\cite{afouras2020self}  & Flickr-10k  & 0.546 & 0.504 \\
LVS~\cite{chen2021localizing}     & Flickr-10k & 0.582 &  0.525 \\ 
{\bf Ours}  &  Flickr-10k & $\textbf{ 0.755}$    & $\textbf{0.588}$   \\
\midrule
Attention~\cite{senocak2018learning}  & Flickr-144k & $0.660$ & $0.558$ \\
DMC~\cite{hu2019deep} & Flickr-144k & $0.671$ & $0.568$ \\
LVS~\cite{chen2021localizing}  & Flickr-144k & $0.699$ & $0.573$ \\
HPS~\cite{senocak2022learning} & Flickr-144k & 0.762 &  0.597 \\
SSPL~\cite{song2022sspl} & Flickr-144k & 0.759 & 0.610 \\
{\bf Ours}  &  Flickr-144k  & $\textbf{0.815}$ & $\textbf{0.611}$ \\
\bottomrule
\end{tabular}
\end{table}

\subsubsection{\bf Ablation Study}
\label{sec:ablation}
In this section, we conduct thorough ablation studies on VGG-SS test, to validate the effectiveness of each component. The results are reported in Table~\ref{tab:ablation}. 
To facilitate comparisons, model-A is set as the baseline with only contrastive loss $\mathcal{L}_{\text{cl}}$ applied, which shares similar setting as LVS~\cite{chen2021localizing}.

\paragraph{\bf Effectiveness of aggressive augmentations.}
When comparing with the baseline, 
model-B~(only appearance augmentation) and model-C~(both appearance and geometrical augmentations) have clearly shown superior performance,
about 3\% cIoU, demonstrating the effectiveness of visual augmentations.
Additionally, when adding audio augmentations~(model-D), 
We observe further performance boost~(about 4.5\% cIoU over baseline).

\paragraph{\bf  Effectiveness of audio mixing.}
On the one hand, comparing model-D and model-E, the proposed audio mixing also brings tiny performance boost. On the other hand, we do observe its benefits for mitigating overfitting issue, as demonstrated in Figure~\ref{fig:cama} of \ref{sec:audmix} in the appendix.
For the model without leveraging audio mixing transformations, the validation cIoU tends to decrease after 40 Epochs, which is a typical performance degradation caused by severe overfitting.
For the model with the audio mixing transformation, the validation loss is constantly decreasing, showing that the overfitting issue is well solved.
In conclusion, our proposed audio mixing transformation can slightly improves localisation performance, as well as preventing the model from overfitting.

\begin{table}[t]
\setlength{\tabcolsep}{10pt}
\caption{Comparisons on the VGG-SS and Flickr-SoundNet test sets. Note that, all models are trained on VGG-Sound 144k. For VGG-SS, our method significantly surpasses previous state-of-the-art models; when evaluating on Flickr-SoundNet test set, our method still performs the best, revealing strong \textit{generalisation} across different datasets.} 
\label{table:quan-vgg}
\begin{center}
\begin{tabular}{lcccc}
\toprule & \multicolumn{2}{c}{ VGG-SS } & \multicolumn{2}{c}{ Flickr-SoundNet } \\
Method & cIoU & AUC & cIoU & AUC \\
\hline
Attention~\cite{senocak2018learning}  & $0.185$ & $0.302$ & $0.660$ & $0.558$ \\
AVO~\cite{afouras2020self} & $0.297$ & $0.357$ & $-$ & $-$ \\
SSPL~\cite{song2022sspl} &  0.339  & 0.380  & 0.767     & 0.605  \\
LVS~\cite{chen2021localizing}  & 0.344 & 0.382 & 0.719 & 0.582 \\
HPS~\cite{senocak2022learning} & 0.346 & 0.380 & 0.768 & 0.592 \\ \midrule
{\bf Ours} & $\textbf{0.386}$ & $\textbf{0.396}$ & $\textbf{0.795}$ & $\textbf{0.612}$ \\
\bottomrule
\end{tabular}
\end{center}
\end{table}

\paragraph{\bf  Effectiveness of geometrical consistency.}
When training model-F with geometrical consistency, 
our best model achieves the best performance, about 6\% cIoU over the baseline model.

\paragraph{\bf  Summary.} 
As shown in Table~\ref{tab:ablation},
all the components including various data augmentation,
{\em e.g.}~appearance and geometrical ones on visual frames, masking, and audio mixing,
are all critical to boosting performance on self-supervised sound source localisation.
Additionally, by further enforcing the audio-visual representation to be equivariant, 
the proposed framework has achieved the best performance.

\subsubsection{\bf Compare with State-of-the-Art}
Here, we compare with the existing methods on the task of sound source localisation,
including: Attention~\cite{senocak2018learning}, AVO~\cite{afouras2020self}, DMC~\cite{hu2019deep}, HPS~\cite{senocak2022learning}, SSPL~\cite{song2022sspl}, CoarseToFine~\cite{qian2020multiple}, and LVS~\cite{chen2021localizing}.

\paragraph{\bf Quantitative Results on Flickr-SoundNet.}
In Table~\ref{table:quan-flickr}, 
we present the comparisons between various approaches on Flickr-SoundNet test set.
Here, we train the model on two training sets, namely, Flickr-10k and Flickr-144k subsets. 
Experimentally, our proposed method outperforms all existing methods by a large margin.
Note that, some methods use additional data or information, e.g., Attention~\citep{senocak2018learning} uses 2796 bounding box annotated audio-image pairs as supervision. 
CoarseToFine~\cite{qian2020multiple} exploits a pretrained object detector to obtain pseudo category labels. 
In contrast, our proposed model is trained from scratch.
Moreover, it can be seen that our model trained on 10k subset performs even better than LVS trained on 144k subset,
that is to say, we achieve superior results with less than 1/14 of training data that the counterpart method~\cite{chen2021localizing} requires, demonstrating the high data-efficiency of our proposed framework.

\paragraph{\bf  Quantitative Results on VGG-SS}
Following~\cite{chen2021localizing}, 
we here train the model on the VGGSound-144k training split, 
but make comparisons between various approaches on the VGG-SS and Flickr-SoundNet test sets,
as shown in Table~\ref{table:quan-vgg}.
On VGG-SS test set, 
our framework surpasses the previous state-of-the-art model~\cite{senocak2022learning} by a noticeable margin.
In addition, when evaluating on Flickr-SoundNet test set, 
our method also maintains its top position, 
revealing strong \textit{generalisation} across different datasets.

\paragraph{\bf Open Set Sound Localisation on VGG-SS}
Following the evaluation protocol in LVS~\cite{chen2021localizing}, 
in this section, we also show the sound localisation results in an open set scenario, 
where models are trained with 110 heard categories in VGGSound, 
and then evaluated on 110 heard and 110 unheard categories separately in the test set.
As shown in Table~\ref{tab:open-loca}, 
both approaches have experienced performance drop on unheard categories,
however, our proposed model still maintains high localisation accuracy in this open set evaluation.

\begin{table}[t]
\setlength{\tabcolsep}{13pt}
\caption{Results for open set sound localisation. All models are trained on 70k samples from 110 object categories in VGGSound, and evaluated on 110 heard categories and 110 unheard categories. Our method shows strong performance.}
\label{tab:open-loca}
\begin{tabular}{clcc}
\toprule
\text { Test class } & \text { Method } & \text { CIoU } & \text { AUC } \\
\hline 
\multirow{2}{*}{\text { Heard 110 }} & \text { LVS~\cite{chen2021localizing}  } & 0.289 & 0.362 \\
& \text { Ours } & \textbf{0.390} & \textbf{0.403} \\
\hline 
\multirow{2}{*}{\text { Unheard 110 }} & \text { LVS~\cite{chen2021localizing} } & 0.263 & 0.347 \\
& \text { Ours } & \textbf{0.365} & \textbf{0.386} \\
\bottomrule
\end{tabular}
\end{table}

\begin{figure*}[t]
\centering
\includegraphics[width=1.0\linewidth]{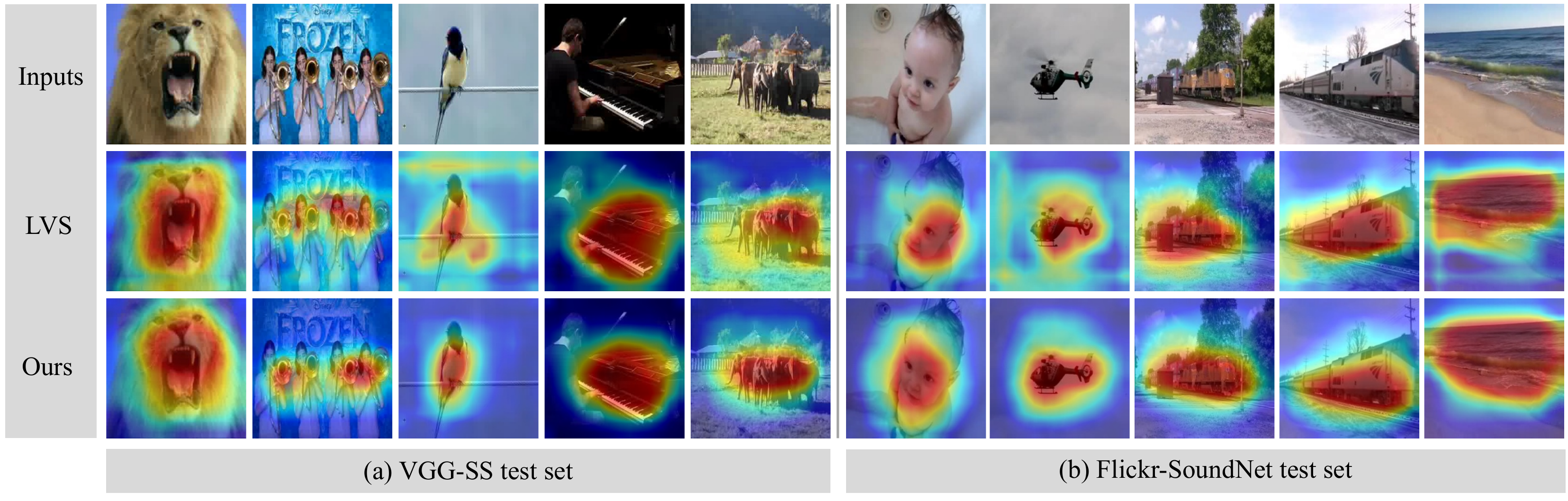}
\caption{Qualitative results on VGG-SS and Flickr-SoundNet test sets for visual sound localisation. 
LVS~\cite{chen2021localizing}, as the state-of-the-art competitor, is chosen for comparison.
The models are trained on Flickr-144k and VGGSound-144k datasets respectively.
Our method localises sounding objects more accurately than LVS, especially for small-size objects.}
\label{fig:qual-compare}
\vspace{0.2cm}
\end{figure*}

\begin{table}[t]
\setlength{\tabcolsep}{6pt}
\caption{Results for audio retrieval. For fair comparisons, all models adopt the ResNet-18 backbone. We here use Accuracy (A@5, A@10) and Precision (P@1, P@5) as metrics. Our learned audio representations are powerful and sometimes comparable to full supervision.}
\label{tab:retri}
\begin{tabular}{lcrrrr}
\toprule
Method & Supervision  & A@5 & A@10 & P@1 & P@5  \\ \midrule
Random & No    & 20.10  & 28.06  &     13.88   & 6.06 \\
VGG-H~\cite{chen2020vggsound} & Full & \textbf{42.07} & \textbf{45.27} & 58.69  & \textbf{27.63}   \\  \midrule
LVS~\cite{chen2021localizing}  & Self  & 26.01 & 33.67 & 21.17  & 9.37  \\ 
{\bf Ours} & Self &   41.15  &  44.19  &\textbf{60.19}   & 27.55    \\ \bottomrule
\end{tabular}
\vspace{-0.2cm}
\end{table}

\subsubsection{\bf Qualitative Results.} 
In Figure~\ref{fig:qual-compare}, we show some qualitative comparisons between LVS~\cite{chen2021localizing} and our proposed method on Flickr-Sound test set and VGG-SS test set.
As can be observed, our model generally produces more accurate localisation results than LVS, in two aspects:
1) our predictions tend to be more complete and highly consistent with the shape of the sounding objects, that means, a more precise prediction on the object boundaries, 
while LVS only localises the parts of objects.
2) our localisation more focuses on the foreground sounding objects, regardless of the background or silent distracting objects; while the localisations of LVS are sometimes scattered even in the clean background, {\em e.g.}, the $third$ column in subplot (a) and the $second$ and $third$ column in subplot (b).

\begin{figure*}[t]
{
\centering
\begin{subfigure}[b]{0.485\textwidth}
    \centering
    \includegraphics[width=\textwidth]{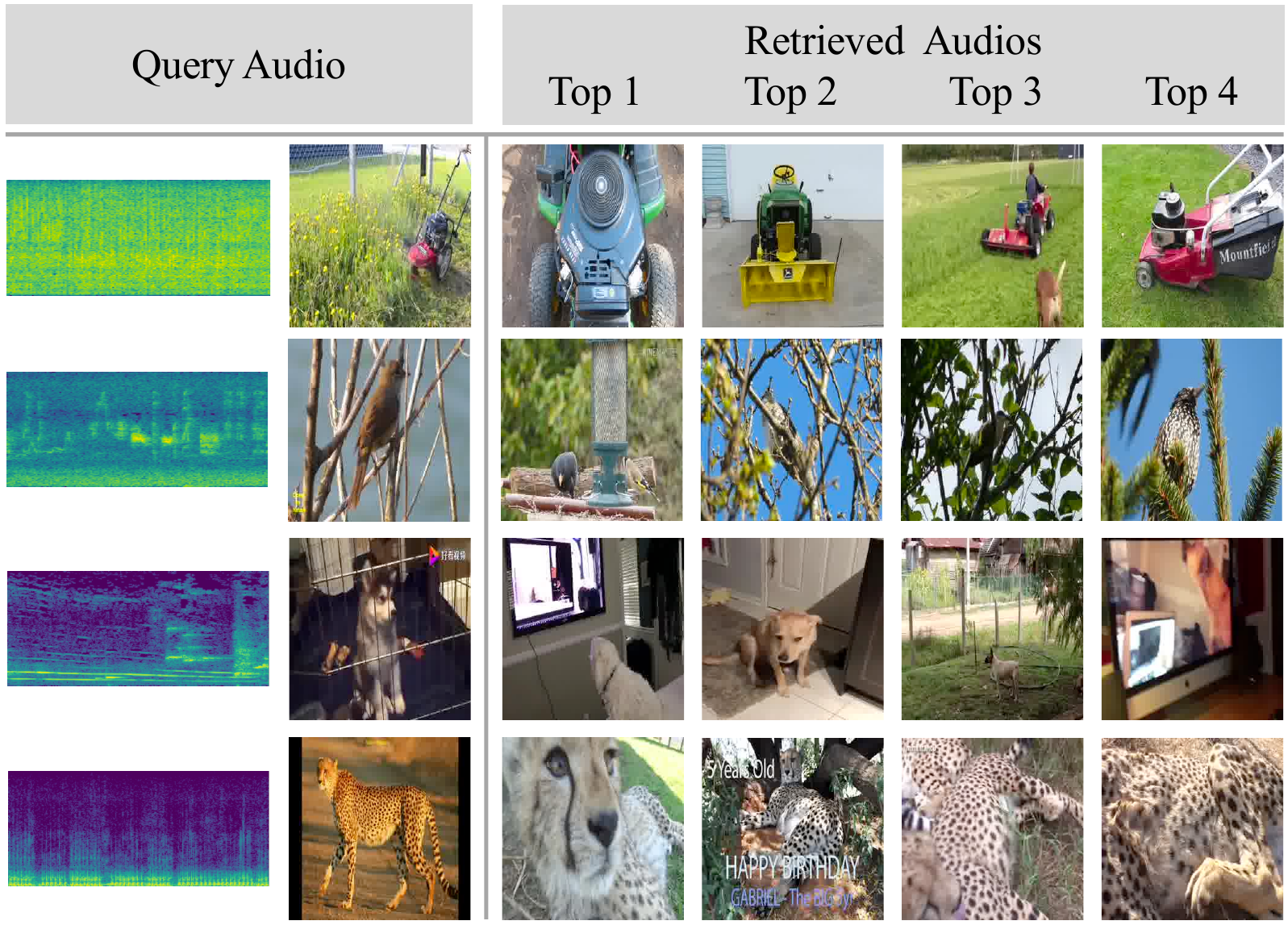}
    \caption{Audio Retrieval}
    \end{subfigure}
    \hfill
\begin{subfigure}[b]{0.485\textwidth}
    \centering
    \includegraphics[width=\textwidth] {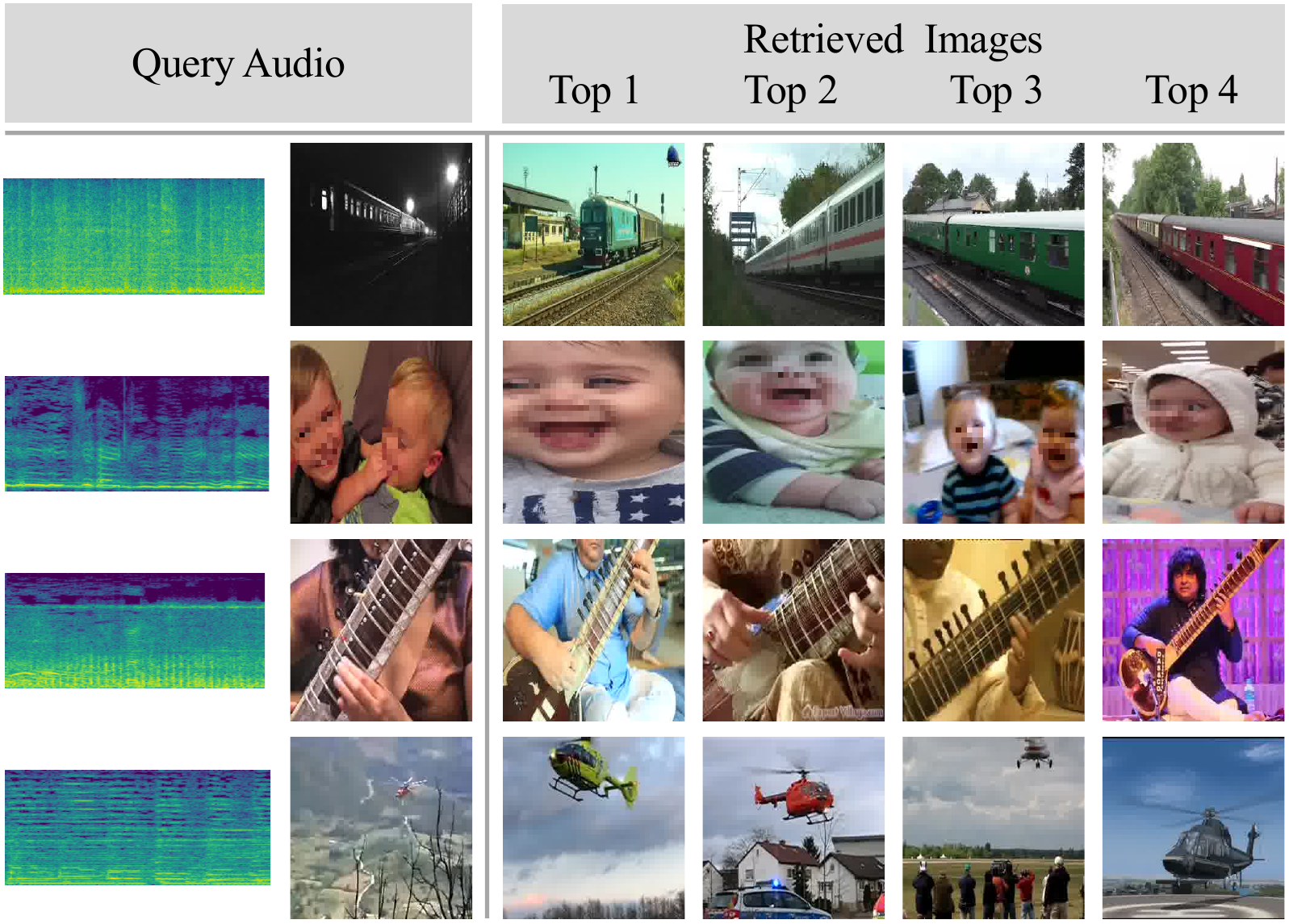}
    \caption{Cross-modal Retrieval}
    \end{subfigure}
\vspace{-0.2cm}
\caption{Qualitative results on retrieval tasks. (a) Audio retrieval, which retrieves semantic-identical audios with the query audio. 
(b) Audio image cross-modal retrieval, which we use the audio as query to retrieve images. The results show our model can accurately retrieve samples with close semantics, indicating our framework has learnt powerful multi-modal representation.
}
\vspace{0.1cm}
\label{fig:vis-retri}
}
\end{figure*}

\subsection{Audio Retrieval} 
To further investigate the quality our our learned audio representation, 
we evaluate the methods on audio retrieval task. 

\subsubsection{\bf Benchmarks}\label{sec:aud-ret}
Due to the lack of unified benchmarks, 
we first divide the VGGSound dataset into train-val set and test set, 
with categories being disjoint.
The former is for training and validation, 
while the latter consisting of unseen categories is for evaluation. 
In detail, the train-val set spans over 274 categories with 169923 samples, 
we randomly sample a 144k subset for training and the rest as the validation set. 
The test set has 35 object categories covering 20304 samples.

\begin{table}[t]
\setlength{\tabcolsep}{6pt}
\caption{Results for audio-image cross-modal retrieval. We report Accuracy (A@5, A@10) and Precision (P@1, P@5). Our model has shown impressive retrieval performance, implying the strong multi-modal representation extraction abilities of our self-supervised models.}
\label{tab:ret-audimg}
\begin{tabular}{lcrrrr}
\toprule
Method & Train Category & A@5 & A@10 & P@1 & P@5  \\
\midrule
Random & 0    & 4.44  & 8.01  &     1.35   & 1.61 \\
LVS~\cite{chen2021localizing}  & All  & 11.00 & 16.26  & 10.48  & 4.93  \\ \midrule
Ours & 110 & 22.44 & 27.72 & 25.50  &  12.34       \\
Ours & All & \textbf{31.67} & \textbf{35.81}  &\textbf{40.91}   & \textbf{19.52}    \\
\bottomrule
\end{tabular}
\end{table}

\subsubsection{\bf Metrics}
We use two standard metrics: \textit{accuracy} and \textit{precision}.
For Top-K accuracy (A@K), as long as the K results contain at least one item of the same category as the query audio,  the retrieval is regarded as correct.
Precision (P@K) is the percentage of the top-K retrieved items of the same category with query audio.

\subsubsection{\bf Baselines}
\label{subsubsec:baselines}
Here, we compare the retrieval results with the following models:
1) Random: the model weights are randomly initialized without training.
2) VGG-H: the model is trained with ground-truth category supervision on the training set, 
as has been done in ~\cite{chen2020vggsound},
3) LVS: a recent state-of-the-art model trained for visual sound localisation~\cite{chen2021localizing}.
4) Ours: our Siamese framework trained on self-supervised visual sound localisation.
For fair comparisons, all models use the ResNet-18 backbone as the audio encoder.

\subsubsection{\bf Retrieval Detail.}
For each query audio in the test set, 
we extract the \textit{512-D} feature with the audio encoder from different models, 
{\em e.g.}~baselines and our model; 
we then calculate the cosine similarity between the query audio and all the rest samples; finally,  we rank the similarity in a descending order, and output the top-K retrieved audios.

\subsubsection{\bf Results.}
We report the results in Table~\ref{tab:retri},
as can be observed,  our self-supervised model significantly outperforms the random and LVS baselines and even demonstrate comparable results to the fully-supervised model, 
{\em i.e.} (VGG-H). 
In Figure~\ref{fig:vis-retri} (a), 
we qualitatively show some audio retrieval results in the form of paired video frames. 
Our model can correctly retrieve samples with close semantics, 
which can potentially be used as auxiliary evidence for video retrieval applications.

\subsection{ Cross-modal Retrieval}
We also evaluate an audio-image cross-modal retrieval task to evaluate the learned cross-modal representations.

\subsubsection{\bf Benchmark.} Similar to Section~\ref{sec:aud-ret}, we obtain the train set and test set from VGGSound dataset.
The test set has 20304 samples spanning 35 categories which are the same as audio retrieval.
The train sets have two versions which both have 144k samples.
The difference is one train set covers all categories while the other train set has 110 categories which are disjoint with test set.

\subsubsection{\bf Metrics.} 
Similar to the audio retrieval task, we also report Top-K accuracy (A@K) and Top-K precision (P@K).

\subsubsection{\bf Baselines.}
We compare the retrieval results with the following models: 1) Random 2) LVS~\cite{chen2021localizing} 3) Ours.
For fair comparisons, all models employ the ResNet-18 backbone as audio and image encoders.
See Section~\ref{subsubsec:baselines} for more details.

\subsubsection{\bf Retrieval Details.}
For each query audio in the test set, we extract \textit{512-D} feature with the audio encoder from different models. For all images to be retrieved in the dataset, we extract the visual features from the visual encoder and spatially pool them into \textit{512-D} vector.
Then we compute the cosine similarity between the query audio and the image samples to be retrieved.
Finally, we rank the similarity in descending order and check the category labels from top-K retrieved images.

\subsubsection{\bf Results.} We report the cross-modal retrieval results in Table~\ref{tab:ret-audimg}.
Comparing with baselines, our representations from self-supervised sound localiser achieve impressive cross-modality retrieval performances, without any finetuning.
We also qualitatively show the results in~Figure~\ref{fig:vis-retri} (b).
The quantitative and qualitative results show that the various transformations in the proposed sound localisation framework have enabled the audio and visual encoders very strong representation abilities.
As a result, our self-supervised framework is remarkably effective for sound source localisation as well as multi-modal retrieval tasks.

\section{Conclusion}
This paper has presented a self-supervised framework for sound source localisation, by fully exploiting various transformations.
The motivation is that appearance transformations and geometrical transformations on image-audio pairs are coming with two implicit but significant properties: \textit{invariance} and \textit{equivariance}.
Invariance refers that the audio-image correspondences are invariant to data transformations; while equivariance denotes the localisation results are equivariant to the geometrical transformations that applied to input images.
Combining these, we propose Siamese networks with dual branches, each branch accepts input data with different transformations on both modalities. 
Thanks to the two properties, the framework is trained in a fully self-supervised way.
Experiments demonstrate our method significantly outperforms current methods in visual sound localisation. Additionally, we also evaluate audio retrieval and cross-modal retrieval tasks, to show the learned powerful multi-modal representations.
Finally, we perform thorough ablation studies to verify the effectiveness of each component in the framework.

\begin{acks}
This work is supported by the National Key R\&D Program of China (No. 2020YFB1406801), 111 plan (No. BP0719010),  and STCSM (No. 18DZ2270700, No. 21DZ1100100), and State Key Laboratory of UHD Video and Audio Production and Presentation.
\end{acks}

\bibliographystyle{ACM-Reference-Format}
\balance
\bibliography{main}


\clearpage
\appendix

\section{Appendix}

\subsection{Data Transformations}

\begin{figure*}[htb]
\centering
\includegraphics[width=\linewidth]{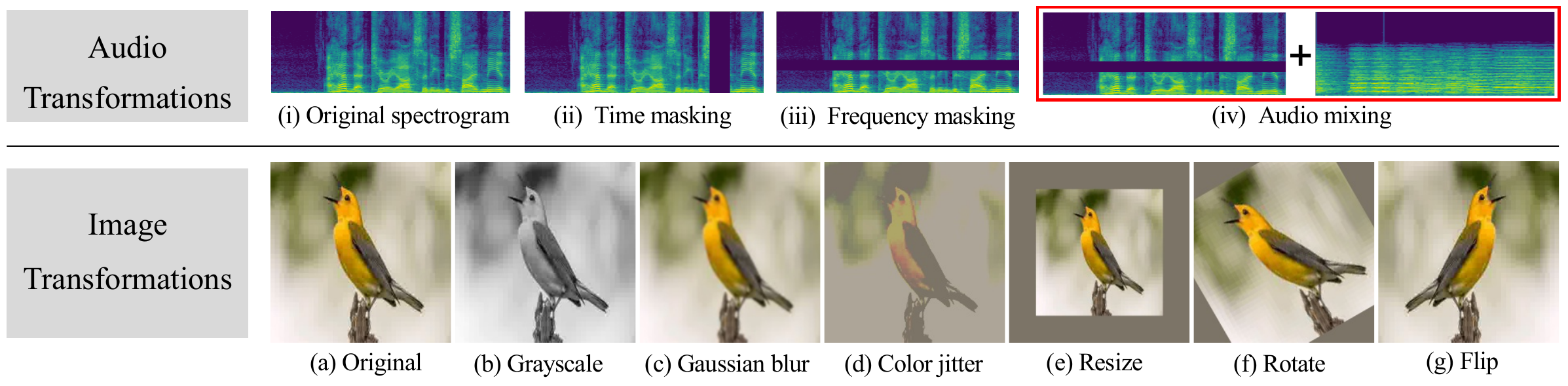}
\vspace{-0.7cm}
\caption{Transformations adopted in the proposed framework. For the visual domain, we explore two types of image transformations: appearance transformation $\mathcal{T}_{app}$ (b-d) and geometrical transformation $\mathcal{T}_{app}$ (e-g); for the audial domain, we apply three effective transformations (ii-iv), which are denoted as $\mathcal{T}_{aud}$.}
\label{fig:transformation}
\end{figure*}

We deploy various data transformation on the input data including audios and video frames.
These transformations are visualized in Figure~\ref{fig:transformation}. 

In the training stage, the audios are mixed with semantic-identical audio samples and then randomly masked with two strategies, namely time masking and frequency masking.
The masking probabilities of two masking strategies are both 0.8.

For the transformations on video frames $\mathcal{T}_{vis}$: the color jitter randomly changes the brightness, contrast, saturation and hue of the image.
And the strength for changing the four factors is the tuple (0.4, 0.4, 0.4, 0.1), each element corresponding to one factor respectively; the application probability is 0.8.
The grayscale transformation is applied with the probability of 0.2.
For the Gaussian blur, the standard deviation for creating the blurring kernel is uniformly sample from  [0.1, 2.0 ]; the application probability is 0.5.
For the geometrical transformations $\mathcal{T}_{geo}$: the resize factor is 0.5 with the probability of 0.5. 
The max rotation degrees is 30.
The horizontal flip operation is deployed with the probability of 0.5.

\subsection{Audio Mixing Transformation}\label{sec:audmix}
\begin{figure*}[htb]
\includegraphics[width=0.55\textwidth]{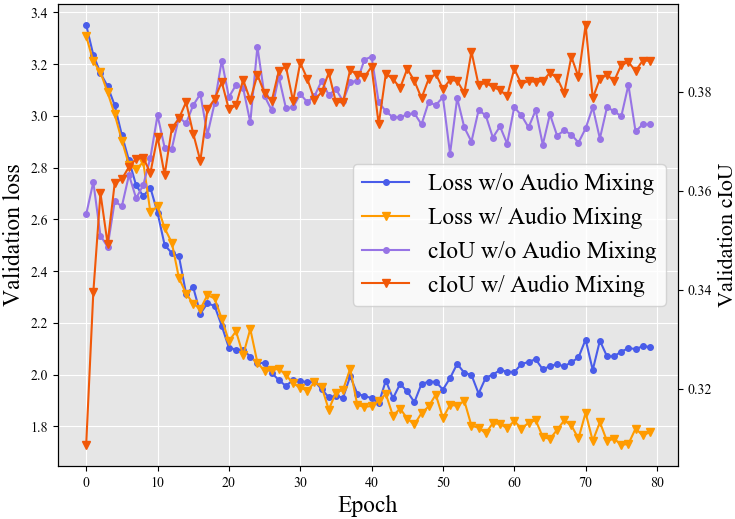}
\caption{Validation curves with or without the audio mixing transformation on VGGSound-144k. Such audio mixing transformation can bring tiny performance boost, and prevent the model from overfitting.}
\label{fig:cama}
\end{figure*}

The validation curves of VGGSound-144k with or without audio mixing are shown in Fig~\ref{fig:cama}. 
As illustrated, the audio mixing transformation can bring performance gains by preventing the model for overfitting.

\end{document}